\documentclass[a4paper]{article}
\usepackage[utf8]{inputenc}
\usepackage[english]{babel}
\usepackage{amsmath}
\usepackage{amsfonts}
\usepackage{float}
\usepackage{amsthm}
\usepackage{amssymb}
\usepackage{graphicx}
\newcommand{\sigm}{\mathop{\mathrm{sigm}}\nolimits}
\newcommand{\Ber}{\mathop{\mathrm{Ber}}\nolimits}
\theoremstyle{definition}
\newtheorem*{example}{Example}
\newcommand{\V}[1]{\mathbf{#1}}
\title{Feature and Variable Selection in Classification}
\author{Aaron Karper}

\begin{document}
\maketitle
\begin{abstract}
The amount of information in the form of features and variables available to machine learning algorithms is ever increasing. This can lead to classifiers that are prone to overfitting in high dimensions, high dimensional models do not lend themselves to interpretable results, and the CPU and memory resources necessary to run on high-dimensional datasets severly limit the applications of the approaches.

Variable and feature selection aim to remedy this by finding a subset of features that in some way captures the information provided best.

In this paper we present the general methodology and highlight some specific approaches.
\end{abstract}
\section{Introduction}
As machine learning as a field develops, it becomes clear
that the issue of finding good features is often more
difficult than the task of using the features to create a
classification model. Often more features are available than 
can reasonably be expected to be used, because using too many features
can lead to overfitting, hinders the interpretability, and
is computationally expensive.

\subsection{Overfitting}
One of the reasons why more features can actually hinder
accuracy is that the more features we have, the less can
we depend on measures of distance that many classifiers (e.g. 
SVM, linear regression, k-means, gaussian mixture models, \dots) require. This is known as the curse of dimensionality.

Accuracy might also be lost, because we are prone to overfit
the model if it incorporates all the features.
\begin{example}
	In a study of genetic cause of cancer, we might end up
	with 15	participants with cancer and 15 without. Each
	participant has 21'000 gene expressions. If we
	assume that any number of genes in combination can 
	cause cancer, even if we underestimate the number of
	possible genomes by assuming the expressions to be
	binary, we end with $2^{21'000}$ possible models.
	
	In this huge number of possible models, there is bound to
	be one arbitrarily complex that fits the observation
	perfectly, but has little to no predictive power \cite[Chapter 18,  Noise and Overfitting]{russell1995artificial}.
	Would we in some way limit the complexity of the model
	we fit, for example by discarding nearly all possible
	variables, we would attain better generalisation.
\end{example}

\subsection{Interpretability}
If we take a classification task and want to gain some
information from the trained model, model complexity can
hinder any insights. If we take up the gene example, a small
model might actually show what proteins (produced by the 
culprit genes) cause the cancer and this might lead to a treatment.

\subsection{Computational complexity}
Often the solution to a problem needs to fulfil certain time
constraints. If a robot takes more than a second to classify
a ball flying at it, it will not be able to catch it. If
the problem is of a lower dimensionality, the computational
complexity goes dows as well.

Sometimes this is only relevant for the prediciton phase of
the learner, but if the training is too complex, it might
become infeasible.

\subsection{Previous work}
This article is based on the work of \cite{guyon2003introduction},
which gives a broad introduction to
feature selection and creation, but as ten years passed, the 
state-of-the-art moved on. 

The relevance of feature selection can be seen in \cite{zhou2005gene}, where
gene mutations of cancer patients are analysed and feature selection
is used to conclude the mutations responsible.

In \cite{torresani2008nonrigid}, the manifold of human poses is modelled
using a dimensionality reduction technique, which will presented here
in short.

Kevin Murphy gives an overview of modern techniques and their justification
in \cite[p. 86ff]{murphy2012machine}

\subsection{Structure}

In this paper we will first discuss the conclusions
of Guyon and Elisseeff about the general approaches taken in
feature selection in section~\ref{sec:classes}, discuss
the creation of new features in section~\ref{sec:creation}, and
the ways to validate the model in section~\ref{sec:validate}.
Then we will continue by showing some more recent developments in the field in section~\ref{sec:examples}.

\section{Classes of methods} \label{sec:classes}
In \cite{guyon2003introduction}, the authors identify four approaches
to feature selection, each of which with its own strengths and
weaknesses:
\begin{description}
	\item[Ranking] orders the features according to some score.
	\item[Filters] build a feature set according to some heuristic.
	\item[Wrappers] build a feature set according to the predictive power of the classifier
	\item[Embedded methods] learn the classification model and the feature selection at the same time.
\end{description}
If the task is to predict as accurately as
possible, an algorithm that has a safeguard against
overfitting might be better than ranking. If a pipeline
scenario is considered, something that treats the following
phases as blackbox would be more useful. If even the time
to reduce the dimensionality is valuable, a ranking would
help.
\subsection{Ranking}
Variables get a score in the ranking approach and the top
$n$ variables are selected. This has the advantage that
$n$ is simple to control and that the selection runs in
linear time.

\begin{example}
In \cite{zhou2005gene}, the authors try to find a discriminative
subset of genes to find out whether a tumor is malignant or
benign\footnote{Acutally they classify the tumors into 3 to 5 classes.}.
 In order to prune the feature base, they rank
the variables according to the correlation to
the classes and make a preliminary selection, which discards
most of the genes in order to speed up the more sophisticated
procedures to select the top 10 features.
\end{example}

There is an inherrent problem with this approach however, called the xor problem\cite{russell1995artificial}:

\begin{figure}[tb]
\includegraphics[width=\linewidth]{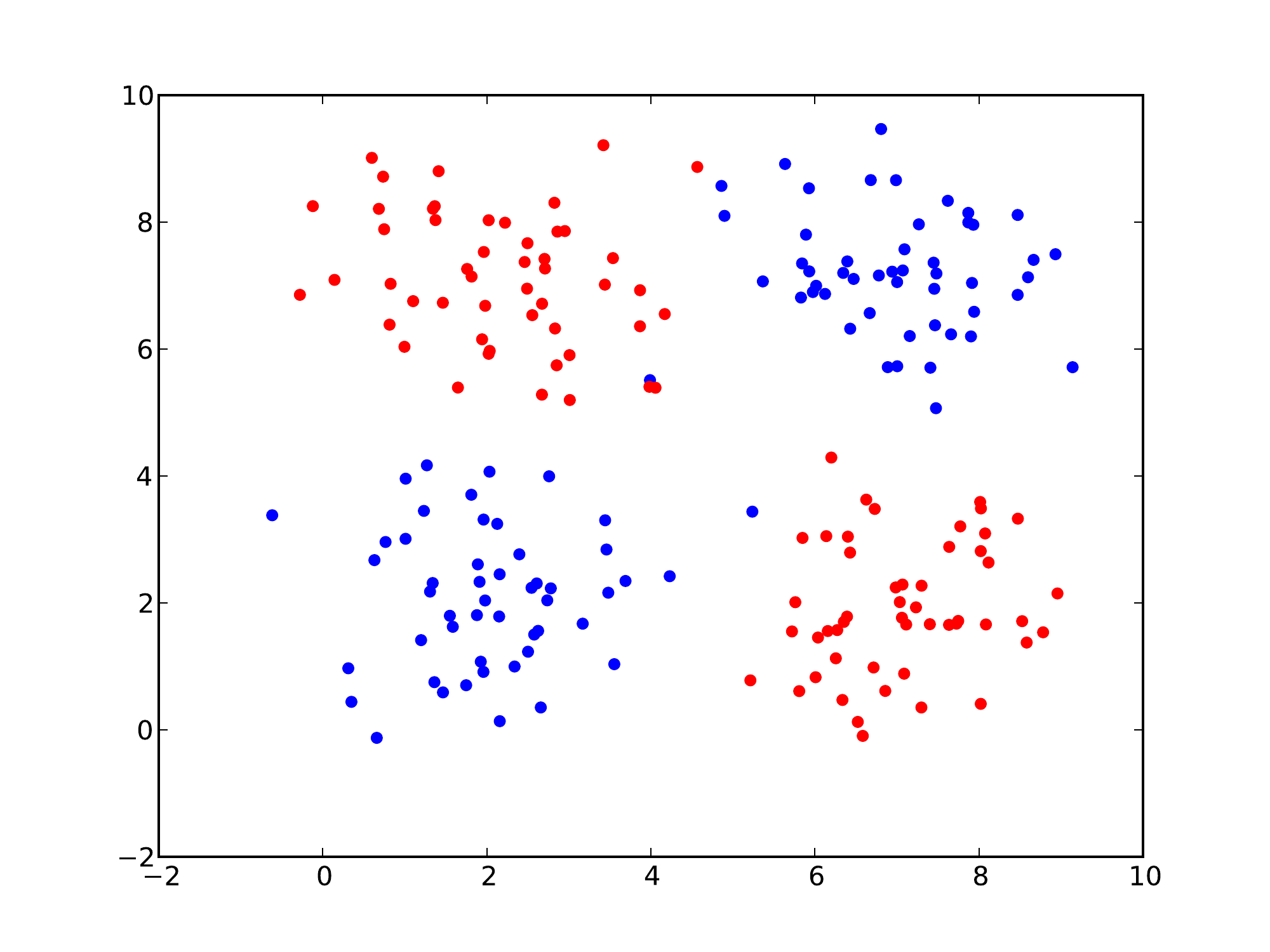}
\label{fig:xor}
\caption{A ranking procedure would find that both features are equally
useless to separate the data and would discard them. If taken together
however the feature would separate the data very well.}
\end{figure}
It implicitly assumes that the features are uncorrelated and gives
poor results if they are not. On figure~\ref{fig:xor}, we have two
variables $X$ and $Y$, with the ground truth roughly being $Z=X>5\; \mathtt{xor}\; Y>5$. Each variable taken separately gives absolutely no information,
 if both variables were selected however,
it would be a perfectly discriminant feature. Since each on its own
is useless, they would not rank high and would probably be discarded
by the ranking procedure, as seen in figure~\ref{fig:xor}.
\begin{example}
Take as an example two genes $X$ and $Y$, so that if one is mutated
the tumor is malignant, which we denote by $M$, but if both mutate, the changes
cancel each other out, so that no tumor grows. Each variable
separately would be useless, because $P(M=true|X=true) = P(Y=false)$, but $P(M=true|X=true,Y=false)=1$ -- 
\end{example}

\subsection{Filters}
While ranking approaches ignore the value that a variable can
have in connection with another, filters select a subset of
features according to some determined criterion. This
criterion is independent of the classifier that is used after
the filtering step. On one hand this allows to only train
the following classifier once, which again might be more cost-effective. On the other hand it also means that only 
some heuristics are available of how well the classifier
will do afterwards.

\begin{figure}[tb]
\includegraphics[width=\linewidth]{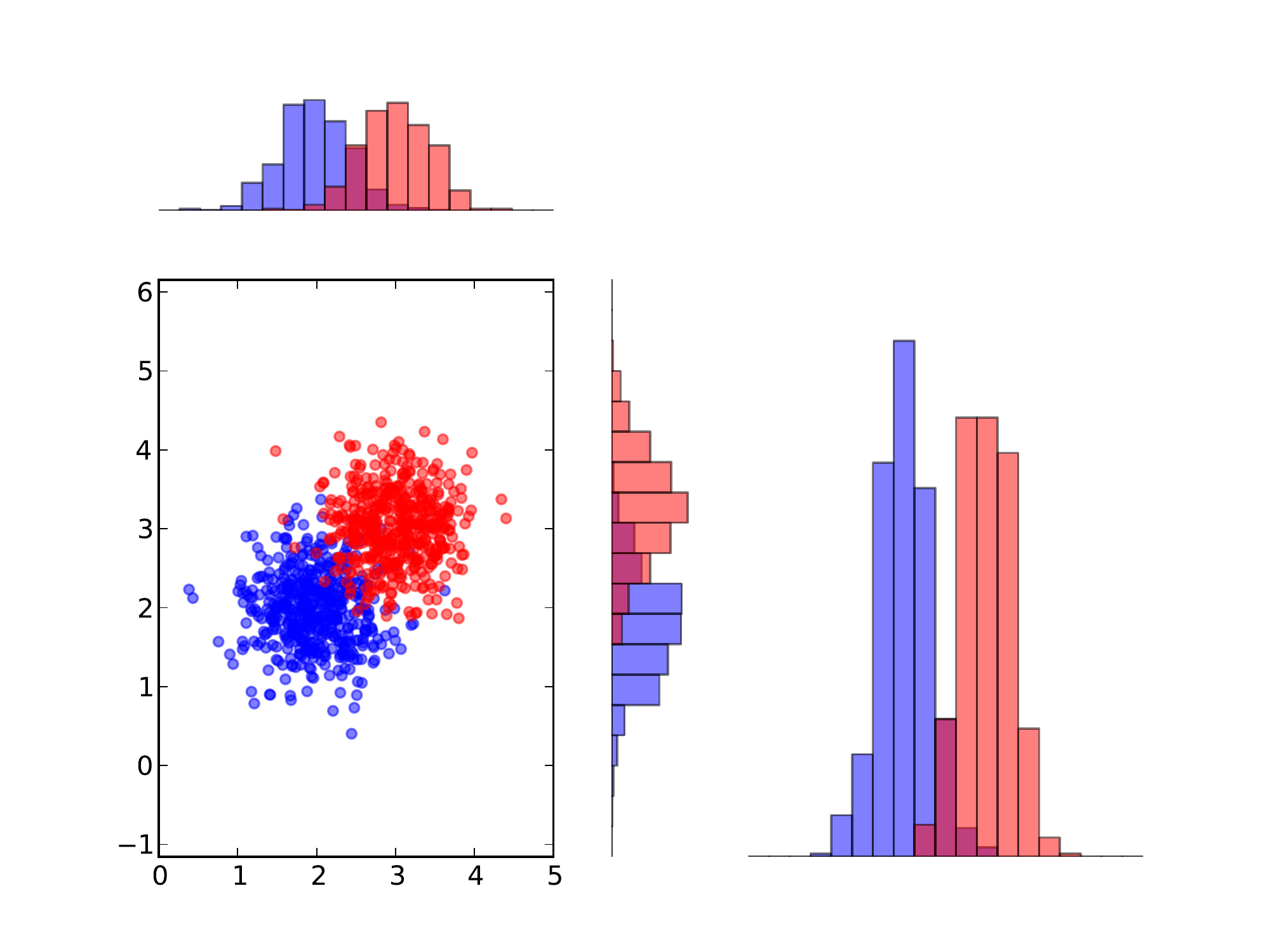}
\label{fig:marginalized}
\caption{Features might be identically distributed, but using both can reduce variance and thus confusion by a factor of $\sqrt{n}$}
\end{figure}

Filtering methods typically try to reduce in-class variance
and to boost inter-class distance. An example of this approach
is a filter that would maximize the correlation between
the variable set and the classification, but minimize
the correlation between the variables themselves. This is
under the heuristic, that variables, that correlate with each
other don't provide much additional information compared to
just taking one of them, which is not necessarily the case, as can be seen on figure~\ref{fig:marginalized}: If the variable is noisy,
a second, correlated variable can be used to get a better
signal, as can be seen in figure \ref{fig:marginalized}.

A problem with the filtering approach is that the performance
of the classifier might not depend as much as we would hope
on the proxy measure that we used to find the subset. In 
this scenario it might be better to assess the accuracy 
of the classifier itself.

\subsection{Wrappers}
\label{ssec:wrappers}
Wrappers allow to look at the classifier as a blackbox and
therefore break the pipeline metaphor. They optimize some
performance measure of the classifier as the objective
function. While this gives superior results to the 
heuristics of filters, it also costs in computation time,
since a classifier needs to be trained each time -- though
shortcuts might be available depending on the classifier
trained.

Wrappers are in large search procedures through feature 
subset space -- the atomic \emph{movements} are to add or to remove
a certain feature. This means that many combinatorical
optimization procedures can be applied, such as simulated
annealing, branch-and-bound, etc. Since the subset space
is $2^N$, for $N$ the number of features, it is not feasible
to perform an exhaustive search, therefore greedy methods
are applied: The start can either be
the full feature set, where we try to reduce the number of
features in an optimal way (\emph{backward elimination}) or
we can start with no features and add them in a smart way 
(\emph{forward selection}). It is also possible to
replace the least predictive feature from the set and
replace it with the most predictive feature from the 
features that were not chosen in this iteration.

\subsection{Embedded}
\label{ssec:embedded}
Wrappers treated classifiers as a black box, therefore 
a combinatorical optimization was necessary with a training
in each step of the search. If the classifier allows feature
selection as a part of the learning step, the learning needs 
to be done only once and often more efficiently.

A simple way that allows this is to optimize in the classifier
not only for the likelihood of the data, but instead for
the posterior probability (MAP) for some prior on the model, 
that makes less complex models more probable.
An example for this can be found in section~\ref{sec:logistic}.

Somewhat similar is SVM with a $\ell_1$ weigth constraint
\footnote{$\ell_p(\mathbf{w})=\|\mathbf{w}\|_p=\sqrt[p]{\sum |w_i|^p}$ }.
The 1:1 exchange means that non-discriminative variables 
will end up with a 0 weight. It is also possible to take 
this a step further by optimizing for the number of variables
directly, since $l_0(w) = \lim_{p\rightarrow 0} l_p(w)$ is exactly
the number of non-zero variables in the vector.

\section{Feature creation} \label{sec:creation}
In the previous chapter the distinction between variables and features
was not necessary, since both could be used as input to the classifier
after feature selection. In this section \emph{features} is the vector
offered to the classifier and \emph{variables} is the vector handed
to the feature creation step, i.e. the \emph{raw inputs} collected.
For much the same reasons that motivated feature selection, feature
creation for a smaller number of features compared to the number of
variables provided.

Essentially the information needs to be compressed
in some way to be stored in fewer variables. Formally this
can be expressed by mapping the high-dimensional space through
the bottleneck, which we hope results in recovering the low dimensional
concepts that created the high-dimensional representation in the
first place. In any case it means that typical features are created, with
a similar intuition to efficient codes in compression:
If a simple feature occurs often, giving it a representation 
will reduce the loss more than representing a less common feature.
In fact, compression algorithms can be seen as a kind of feature
creation\cite{argyriou2008convex}.

This is also related to the idea of manifold learning:
While the variable space is big, the actual space in that the variables
vary is much smaller -- a manifold\footnote{A manifold is the mathematical generalization of a surface or a curve in 3D space: Something smooth that can be mapped from a lower dimensional space.} of hidden variables embedded in the variable space.

\begin{example} In \cite{torresani2008nonrigid} the human body is modelled
as a low dimensional model by \emph{probabilistic principal component analyis}:
It is assumed that the hidden variables are distributed as Gaussians in
a low dimensional space that are then linearly mapped to the high dimensional
space of positions of pixels in an image. This allows them to learn 
\emph{typical} positions that a human body can be in and with that
track body shapes in 3d even if neither the camera, nor the human are fixed.
\end{example}

\section{Validation methods}	\label{sec:validate}
The goal up to this point was to find a simple model, that performs well
on our training set, but we hope that our model will perform well in
data, it has never seen before: minimizing the \emph{generalization error}.
This section is concerned with estimating this error.

A typical approach is \emph{cross-validation}: If we have independent and identically
distributed datapoint, we can split the data and train the model on one
part and measure its performance on the rest. But even if we assume that
the data is identically distributed, it requires very careful curation of
the data to achieve independence:

\begin{example}
Assume that we take a corpus of historical books and segment them. We could
now cross-validate over all pixels, but this would be anything but independent.
If we are able to train our model on half the pixels of a page and check
against the other half, we would naturally perform quite well, since we
are actually able to learn the style of the page. If we split page-wise, we
can learn the specific characteristics of the author. Only if we split
author-wise, we might hope to have a resemblence of independence. 
\end{example}

Another approach is \emph{probing}: instead of modifying the data set
and comparing to other data, we can modify the \emph{feature space}.
We add random variables, that have no predictive power to the
feature set. Now we can measure how well models fare against pure chance\footnote{
This can take the form of a significance test.}. Our performance measure 
is then the signal-to-noise ratio of our model.

\section{Current examples}
\label{sec:examples}
\subsection{Nested subset methods}
\label{sec:nested}
In the nested subset methods the feature subset space is greedily 
examined by estimating the expected gain of adding one 
feature in forward selection or the expected loss of removing 
one feature in backward selection. This estimation is called
the \emph{objective function}. If it is possible to examine 
the objective function for a classifier directly, a better
performance is gained by embedding the search procedure with
it. If that is not possible, training and evaluating the classifier
is necessary in each step.

\begin{example} Consider a model of a linear predictor $p(\V{y}|\V{x})$
with $M$ input variables needing to be pruned to $N$ input variables. 
This can be modeled by asserting that the real variables $x_i^\star$ are taken from 
$\mathbb{R}^N$, but a linear transformation $A\in \mathbb{R}^{N\times M}$
and a noise term $n_i=\mathcal{N}(0, \sigma_x^2)$ is added:
\[ \V{x_i} = A\V{x^\star_i}+n_i\]
In a classification task\footnote{The optimization a free interpretation of \cite{guo2008supervised}}, we can model $y= \Ber(\sigm(\V{w}\cdot \V{x}^\star))$.

This can be seen as a generalisation of PCA\footnote{Principal component analysis reduces the dimensions of the input variables by taking only the directions of the largest eigenvalues.} to the case where the output variable
is taken into account (\cite{west2003bayesian} and \cite{bair2006prediction} develop
the idea). Standard supervised PCA assumes that the output is
distributed as a gaussian distribution, which is a dangerous simplification
in the classification setting\cite{guo2008supervised}.

The procedure iterates over the eigenvectors of the natural parameters
of the joint distribution of the input and the output
and adds them if they show an improvement to the current model in order to
capture the influence of the input to the output optimally. If 
more than $N$ variables are in the set, the one with the least favorable score
is dropped. The algorithms iterates some fixed number of times over all
features, so that hopefully the globally optimal feature subset is found.
\end{example}

\subsection{Logistic regression using model complexity regularisation}
\label{sec:logistic}
In the paper \emph{Gene selection using logistic regressions 
based on AIC, BIC and MDL criteria}\cite{zhou2005gene} by Zhou, Wang, and Dougherty,
the authors describe the
problem of classifying the gene expressions that determine
whether a tumor is part of a certain class (think malign
versus benign). Since the feature vectors are huge 
($\approx 21'000$ genes/dimensions in many expressions) and
therefore the chance of overfitting is high and the domain
requires an interpretable result, they discuss feature selection.

For this, they choose an embedded method, namely a normalized form
of logistic regression, which we will describe in detail here:

Logistic regression can be understood as fitting 
\[ p_\mathbf{w}(\mathbf{x})=\frac{1}{1+e^{\mathbf{w}\cdot \mathbf{x}}}= \sigm(\mathbf{w}\cdot \mathbf{x}) \]
with regard to the separation direction $\mathbf{w}$, so that the 
confidence or in other words the probability 
$p_{\mathbf{w}^\star}(\mathbf{x}_{data})$ is maximal.

This corresponds to the assumption, that the probability of
each class is $p(c|\mathbf{w}) = \Ber(c|\sigm(\mathbf{w}\cdot \mathbf{x}))$
and can easily be extended to incorporate some prior on $\mathbf{w}$,
$p(c|\mathbf{w}) = \Ber(c|\sigm(\mathbf{w}\cdot \mathbf{x}))\, p(w)$ \cite[p. 245]{murphy2012machine}.

The paper discusses the priors of the Akaike information criterion (AIC),
the Bayesian information criterion (BIC) and the minimum descriptor
length (MDL):

\paragraph{AIC} The Akaike information criterion penaltizes degrees
of freedom, the $\ell_0$ norm, so that the function optimized in
the model is $\log L(\mathbf{w}) - \ell_0(\mathbf{w})$. This corresponds
to an exponential distribution for $p(\mathbf{w}) \propto \exp(-\ell_0(\mathbf{w}))$. This can be interpreted as minimizing
the variance of the models, since the variance grows exponentially
in the number of parameters.

\paragraph{BIC} The Bayesian information criterion is similar, but
takes the number of datapoints $N$ into account: 
$p(\mathbf{w}) \propto N^{-\frac{\ell_0(\mathbf{w})}{2}}$. This has an
intuitive interpretation if we assume that a variable is either ignored,
in which case the specific value does not matter, or taken into account, 
in which case the value influences the model. If we assume a uniform
distribution on all such models, the ones that ignore become more
probable, because they accumulate the probability weight of all possible
values.

\paragraph{MDL} The minimum descriptor length is related to the algorithmic
probability and states that the space necessary to store the descriptor
gives the best heuristic on how complex the model is. This only implicitly
causes variable selection. The approximation for this value can be seen in
the paper itself.

Since the fitting is computationally expensive, the authors start with
a simple ranking on the variables to discard all but the best 5'000. 
They then repeatedly fit the respective models and collect the number
of appearances of the variables to rank the best 5, 10, or 15 genes.
This step can be seen as an additional ranking step, but this seems
unnecessary, since the fitted model by construction would already
have selected the best model. Even so they still manage to avoid 
overfitting and finding a viable subset of discriminative variables.

\subsection{Autoencoders as feature creation}
Autoencoders are deep neural networks\footnote{Deep means multiple hidden layers.} that find a fitting information bottleneck (see \ref{sec:creation})
by optimizing for the reconstruction of the signal using the \emph{inverse}
transformation\footnote{A truely inverse transformation is of course not 
possible.}.

\begin{figure}[tb]
\center{\includegraphics[width=.5\linewidth]{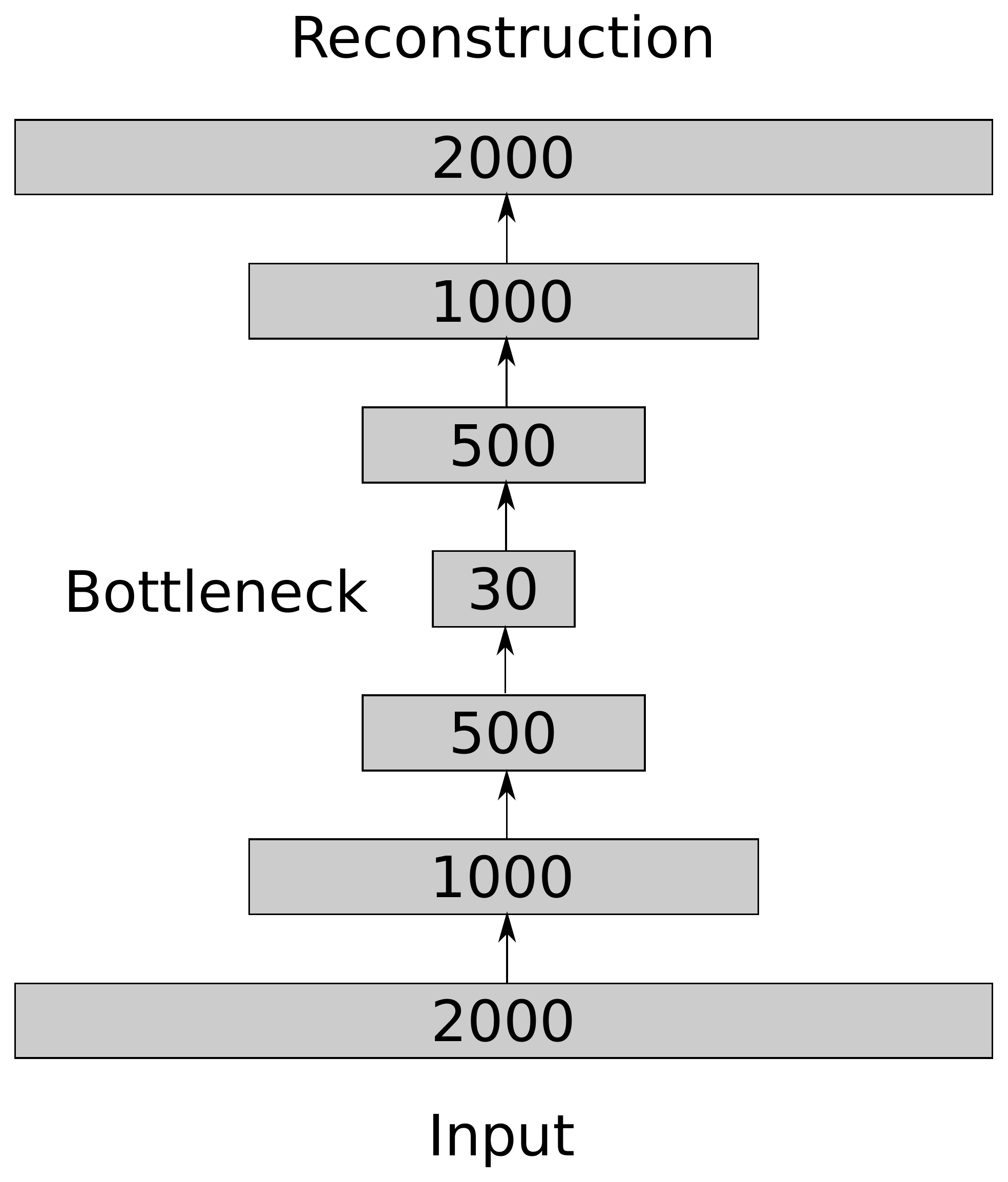}}
\caption{An autoencoder network}
\end{figure}

Deep networks are difficult to train, since they show many local minima,
many of which show poor performance \cite[p. 1000]{murphy2012machine}. 
To get around this, Hinton and Salakhutdinov \cite{hinton2006reducing}
propose pretraining the model as stacked restricted Bolzmann machines
before devising a global optimisation like stochastical gradient descent.

Restricted Bolzmann machines are easy to train and can be understood
as learning a probability distribution of the layer below. Stacking them
means extracting probable distributions of features, somewhat similar to
a distribution of histograms as for example HoG or SIFT being
representative to the visual form of an object.

It has long been speculated that only low-level features could be captured
by such a setup, but \cite{le2011building} show that, given enough resources,
an autoencoder can learn high level concepts like recognizing a cat face
without any supervision on a 1 billion image training set.

The impressive result beats the state of the art in supervised learning
by adding a simple logistic regression on top of the bottleneck layer.
This implies that the features learned by the network capture the concepts
present in the image better than SIFT visual bag of words or other human created features and
that it can learn a variety of concepts in parallel. Further since the
result of the single best neuron is already very discriminative, it
gives evidence for the possibility of a \emph{grandmother neuron} in the
human brain -- a neuron that recognizes exactly one object, in this
case the grandmother. Using this single feature would also take 
feature selection to the extreme, but without the benefit of being
more computationally advantageous.

\subsection{Segmentation in Computer Vision}
A domain that necessarily deals with a huge number of
dimensions is computer vision. Even only considering VGA
images, in which only the actual pixel values are taken into
account gives $480\times 640 = 307'200$ datapoints per image.

For a segmentation task in document analysis, where pixels need 
to be classified into regions like border, image, and 
text, there is more to be taken into account than just the
raw pixel values in order to incorporate spartial information, edges, etc. 
With up to 200 heterogeneous features to consider 
for each pixel, the evaluation would take too long to be useful.

This section differs from the previous two in that instead of
reviewing a ready made solution to a problem, it shows the process
of producing such a solution.

The first thing to consider is whether or not we have a strong
prior of how many features are useful. In the example of cancer
detection, it was known that only a small number of mutation caused
the tumor, so a model with a hundred genes could easily be discarded.
Unfortunately this is not the case for segmentation, because our
features don't have a causal connection to the true segmentation.
Finding good features for segmentation requires finding
a good proxy feature set for the true segmentation.

Next we might consider the loss of missclassification: In a computer 
vision task, pixel missclassifications are to be expected and can
be smoothed over. Computational complexity however can severely
limit the possible applications of an algorithm. As \cite{russell1995artificial}
note, using a bigger dataset can be more advantageous than using
the best algorithm, so we would favour an efficient procedure over a
very accurate one, because it would allow us to train on a bigger training
set. Since the variables are likely to be correlated,
ranking will give bad results. 

Taking this into account, we would consider $L_1$ normalized linear
classifiers, because of the fast classification and training (the
latter due to \cite{yuan2010comparison}, in which linear time training
methods are compared). Taking linear
regression could additionally be advanageous, since its \emph{soft} 
classification would allow for better joining of continuous areas of
the document. 
\section{Discussion and outlook}
Many of the concepts presented in \cite{guyon2003introduction} still apply,
however the examples fall short on statistical justification. Since then
applications for variable and feature selection and feature creation 
were developed, some of which were driven by advances in computing power,
such as high-level feature extraction with autoencoders, others were
motivated by integrating prior assumptions about the sparcity of the model,
such as the usage of probabilistic principal component analysis for 
shape reconstruction.

The goals of variable and feature selection -- avoiding overfitting, interpretability, and computational efficiency -- are in our opinion
problems best tackled by integrating them into the models learned by
the classifier and we expect the embedded approach to be best fit to
ensure an optimal treatment of them. Since many popular and efficient
classifiers, such as support vector machines, linear regression, and
neural networks, can be extended to incorporate such constraints with
relative ease, we expect the usage of ranking, filtering, and wrapping
to be more of a pragmatic first step, before sophisticated learners 
for sparse models are employed. Advances in embedded approaches will
make the performance and accuracy advantages stand out even more.

Feature creation too has seen advances, especially in efficient 
generalisations of the principal component analysis algorithm, 
such as kernel PCA (1998) and supervised extensions. They predominantly
rely on the bayesian formulation of the PCA problem and we expect
this to drive more innovation in the field, as can be seen by the spin-off
of reconstructing a shape from 2D images using a bayesian network as discussed
in \cite{torresani2008nonrigid}. 

\nocite{zhang2011convex,russell1995artificial}

\bibliography{main}

\end{document}